%% file: root.tex
\documentclass[letterpaper, 10 pt, conference]{ieeeconf}  

\IEEEoverridecommandlockouts                              

\makeatletter
\let\NAT@parse\undefined
\makeatother
\input{macros/packages.tex}
\input{macros/notations.tex}

\title{\LARGE \bf
Diffusion Meets Options: Hierarchical Generative Skill Composition for Temporally-Extended Tasks
}

\author{Zeyu~Feng$^{1}$, Hao~Luan$^{1}$, Kevin Yuchen~Ma$^{1}$, and Harold~Soh$^{1,2}$
\thanks{$^{1}$Department of Computer Science, School of Computing, National University of Singapore, Singapore. {\tt\small \{zeyu, haoluan, mayuchen, harold\}@comp.nus.edu.sg}}%
\thanks{$^{2}$Smart Systems Institute, National University of Singapore.}%
}

\begin{document}

\maketitle
\thispagestyle{empty}
\pagestyle{empty}

\begin{abstract}

Safe and successful deployment of robots requires not only the ability to generate complex plans but also the capacity to frequently replan and correct execution errors. This paper addresses the challenge of long-horizon trajectory planning under temporally extended objectives in a receding horizon manner. To this end, we propose \method, a data-driven hierarchical framework that generates and updates plans based on instruction specified by linear temporal logic (LTL). Our method decomposes temporal tasks into chain of options with hierarchical reinforcement learning from offline non-expert datasets. It leverages diffusion models to generate options with low-level actions. We devise a determinantal-guided posterior sampling technique during batch generation, which improves the speed and diversity of diffusion generated options, leading to more efficient querying. Experiments on robot navigation and manipulation tasks demonstrate that \method can generate sequences of trajectories that progressively satisfy the specified formulae for obstacle avoidance and sequential visitation. Demonstration videos are available online at: https://philiptheother.github.io/doppler/.

\end{abstract}

\input{sections/intro.tex}
\input{sections/preliminaries.tex}
\input{sections/method.tex}
\input{sections/relatedwork.tex}
\input{sections/experiments.tex}
\input{sections/conclusion.tex}

\clearpage

\balance
\bibliographystyle{IEEEtran}
\bibliography{IEEEabrv,references}

\end{document}

%% file: macros/packages.tex
\usepackage{algorithm}
\usepackage{algpseudocode}

\usepackage{amsmath}
\usepackage{amstext}
\usepackage{amssymb}
\usepackage{amsfonts}
\usepackage{mathrsfs}
\usepackage{mathtools}
\usepackage{latexsym}
\usepackage{bm}
\usepackage{bbm}
\usepackage{dsfont}
\usepackage{nicefrac}
\usepackage[super]{nth}
\usepackage{wasysym}
\usepackage{pifont}

\usepackage{array}
\usepackage{multirow}
\usepackage{booktabs}
\usepackage{tikz}
\usepackage{xspace}
\usepackage{subfigure}
\usepackage{balance}
\usepackage{threeparttable}

\usepackage[noadjust]{cite}
\usepackage[bookmarks=true]{hyperref}
\usepackage{xcolor}

%% file: macros/notations.tex
\usepackage{mathtools}
\usepackage{amssymb}
\usepackage{amsmath}
\usepackage{tikz}
\usepackage{xspace}

\newcommand{\btau}[1]{\bm{\tau}^{#1}}
\newcommand{\bsigma}{{\bm{\sigma}}}

\newcommand*\diff{\mathop{}\!\mathrm{d}}

\newcommand{\paren}[1]{\left(#1 \right)}
\newcommand{\bracket}[1]{\left[#1 \right]}
\newcommand{\bracelr}[1]{\left\{#1 \right\}}
\newcommand{\expect}[2]{\mathbb{E}_{#1} \left[#2 \right]}
\newcommand{\prob}[2]{p_{#1} \left(#2\right)}
\newcommand{\cprob}[3]{p_{#1} \left(#2 |#3\right)}
\newcommand{\defeq}{\mathrel{\mathop:}=}

\newcommand{\tuple}[1]{\langle{#1}\rangle}

\newcommand{\ltlf}{\textnormal{LTL}_f}

\newcommand{\ltleventually}{\ensuremath{\Diamond}\xspace}
\newcommand{\ltlnext}{\ensuremath{\bigcirc}\xspace}

\newcommand{\true}{\ensuremath{\mathsf{true}}\xspace{}}
\newcommand{\false}{\ensuremath{\mathsf{false}}\xspace{}}
\newcommand{\ltluntil}{\ensuremath{\operatorname{\mathsf{U}}}\xspace{}}

\newcommand{\mprog}{\operatorname{prog}}

\newlength\squareheight
\definecolor{gold}{rgb}{1.0, 0.84, 0.0}

\makeatletter
\algnewcommand{\LineComment}[1]{\Statex \hskip\ALG@thistlm \(\triangleright\) #1}
\makeatother

\newcommand{\ie}{\textit{i.e.}}
\newcommand{\eg}{\textit{e.g.}}

\newcommand{\tabref}[1]{TABLE~\ref{#1}}
\newcommand{\parabf}[1]{\vspace{0.1cm}\noindent\textbf{#1}}

\newcommand{\Dtraj}{\mathcal{D}}
\newcommand{\Dphi}{\mathcal{D}_\Psi}
\newcommand{\Diffusion}[1]{\widehat{p}_\theta(#1|\mathbf{s})}

\newcommand{\diffuser}{\textsc{Diffuser}\xspace}
\newcommand{\diffusionpolicy}{\textsc{Diffusion Policy}\xspace}
\newcommand{\ltldog}{\textsc{LTLDoG}\xspace}
\newcommand{\ltldogr}{\textsc{LTLDoG-R}\xspace}
\newcommand{\ltldogs}{\textsc{LTLDoG-S}\xspace}
\newcommand{\method}{\textsc{Doppler}\xspace}

%% file: sections/intro.tex
\section{Introduction}
\label{sec:intro}

Robots and autonomous agents are increasingly expected to safely perform complex tasks specified by high-level, temporally extended instructions. Linear Temporal Logic (LTL) provides a formal language to specify such tasks, enabling agents to reason about sequences of events over time~\cite{BACCHUS2000123,1570410}. However, planning under LTL constraints poses significant challenges, especially in offline settings where agents must learn from fixed datasets without active exploration. Traditional data-driven policy learning methods struggle with the non-Markovian nature of LTL rewards, and offline RL further complicates the issue due to distributional shifts. 

\begin{figure}
    \centering
    \includegraphics[width=.9\columnwidth]{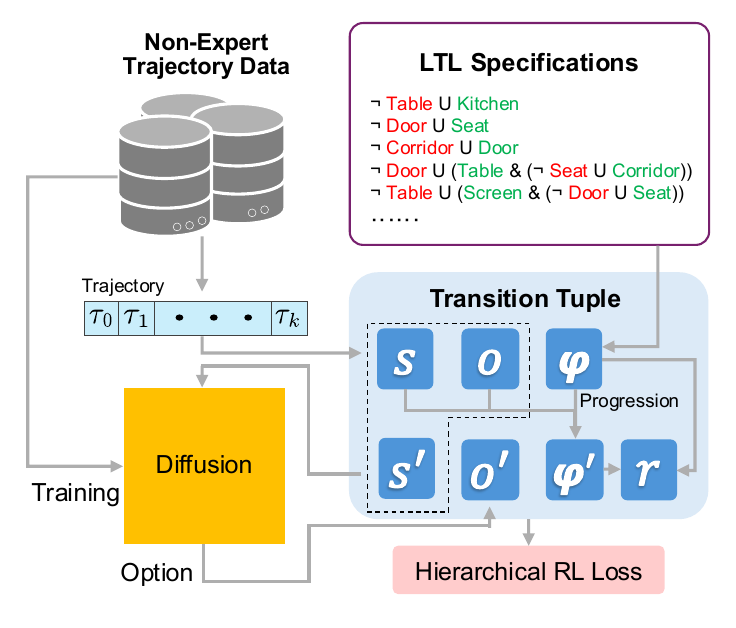}
    \vspace{-15pt}
    \caption{Overview of \method{} framework. The model is trained using non-expert trajectory data and LTL specifications. Trajectories are sampled from the diffusion model to generate options, which are then used to form transition tuples that include the state ($s$), option ($o$), and LTL formula ($\varphi$). These tuples are processed with LTL progression to produce the next state ($s'$), next option ($o'$), updated LTL formula ($\varphi'$), and reward ($r$). The hierarchical RL loss is then computed to guide the learning process.}
    \label{fig:illustration_framework}
    \vspace{-15pt}
\end{figure}

In this work, we propose \textbf{D}iffusion \textbf{O}ption \textbf{P}lanning by \textbf{P}rogressing \textbf{L}TLs for \textbf{E}ffective \textbf{R}eceding-horizon control (\method{}), an offline hierarchical reinforcement learning framework that generates receding horizon trajectories to satisfy given LTL instructions. Our key insight is to leverage diffusion models to represent options within the hierarchical RL framework. By integrating diffusion-based options, we can generate diverse and expressive behaviors while ensuring that the generated trajectories remain within the support of the offline dataset. To address the challenges of option selection and policy regularization in the offline setting, we introduce a diversity-guided sampling approach that promotes exploration of different modes of the data distribution without straying into out-of-distribution actions.

Compared to prior work, the most closely related methods are those that use diffusion models for trajectory planning under temporal constraints~\cite{10637680} and hierarchical diffusion frameworks for subgoal generation~\cite{pmlr-v202-li23ad,ma2024hierarchical}. However, these approaches either lack the ability to perform closed-loop, hierarchical planning or require expert-level demonstrations, limiting their applicability in offline settings with non-expert data. In contrast, \method{} combines hierarchical RL with diffusion-based options to enable closed-loop control under LTL constraints using only offline data. To our knowledge, \method{} is the first work to integrate diffusion models into a hierarchical RL framework for data-driven LTL planning in an offline setting.

Our experiments demonstrate that \method{} significantly outperforms strong baselines on tasks specified by LTL formulas, achieving higher success rates in satisfying temporal specifications. In simulation, \method{} effectively handles complex, long-horizon navigation tasks that prior methods struggle with. On the real robot, \method{} exhibits robust behavior in the presence of control noise and external perturbations, successfully accomplishing LTL tasks where other methods fail.

\method{} represents a step toward enabling autonomous agents to perform complex, temporally extended tasks specified by LTL instructions in offline settings. In summary, this paper makes three key contributions:
\begin{itemize}
    \item A hierarchical RL approach designed for data-driven LTL planning with diffusion-based options;
    \item A diversity-guided sampling approach for diffusion models that enhances option discovery and policy regularization;
    \item Experimental results that validate the effectiveness of receding horizon planning for LTL tasks and demonstrate robustness to runtime deviations.
\end{itemize}
By integrating diffusion models into hierarchical RL and introducing diversity-guided sampling, \method{} enables agents to effectively learn and plan under complex temporal constraints from offline data, paving the way for more capable and reliable autonomous systems.

%% file: sections/preliminaries.tex
\section{Preliminaries}

In this work, we aim to train an agent using an offline dataset to learn policies that generate trajectories satisfying Linear Temporal Logic (LTL) specifications. Our approach employs offline hierarchical reinforcement learning, where options are represented using diffusion models. In this section, we provide a succinct review of LTL, hierarchical reinforcement learning, and diffusion policies.

\subsection{Formal Task Specification via LTL}
\label{sec:pre_ltl}

To unambiguously define the task, we employ Linear Temporal Logic (LTL). LTL is a propositional modal logic with temporal modalities~\cite{4567924} that allows us to formally specify properties or events characterizing the successful completion of a task. These properties and events are drawn from a domain-specific finite set of atomic propositions $\mathcal{P}$. The set of LTL formulae $\Psi$ is recursively defined in Backus-Naur form as follows~\cite{baier0020348,belta2017formal}:
\begin{equation*}
    \varphi \defeq p \;|\; \neg \varphi \;|\; \varphi \wedge \psi \;|\; \ltlnext \varphi \;|\; \varphi \ltluntil \psi,
\end{equation*}
where $p \in \mathcal{P}$ and $\varphi,\psi\in\Psi$. $\neg$ (\textit{negation}) and $\wedge$ (\textit{and}) are Boolean operators while $\ltlnext$ (\textit{next}) and $\ltluntil$ (\textit{until}) are temporal operators. These basic operators can be used to derive other commonly-used operators,
$\true = \varphi \vee \neg\varphi$,
$\varphi \vee \psi = \neg \left(\neg\varphi \wedge \neg\psi\right)$ and
$\ltleventually \varphi = \true \ltluntil \varphi$ (\textit{eventually} $\varphi$).

LTL formulae are evaluated over infinite sequences of truth assignments $\bsigma =\langle \sigma_0, \sigma_1, \sigma_2, \ldots \rangle$, where $\sigma_t\in \{0,1\}^{\vert\mathcal{P}\rvert}$ and $\sigma_{t,p}=1$ iff $p$ is true at time step $t$. $\tuple{\bsigma,t}\models\varphi$ denotes $\bsigma$ \emph{satisfies} $\varphi$ at time $t \geq 0$, which can be initially defined over atomic propositions and then extended to more logical operators according to their semantics. Given a $\sigma$, an LTL $\varphi$ can be \emph{progressed} to reflect which parts of $\varphi$ have been satisfied and which remain to be achieved~\cite{BACCHUS2000123}. The one-step progression of $\varphi$ on $\sigma$ is defined as follows:
\begin{itemize}
    \item $\mprog\paren{\sigma,p} = \true$ if $\sigma_{p}=1$, where $p \in \mathcal{P}$
    \item $\mprog\paren{\sigma,\neg \varphi} = \neg \mprog\paren{\sigma,\varphi}$
    \item $\mprog\paren{\sigma,\varphi\wedge\psi} = \mprog\paren{\sigma,\varphi} \wedge \mprog\paren{\sigma,\psi}$
    \item $\mprog\paren{\sigma,\ltlnext\varphi} = \varphi$
    \item $\mprog\paren{\sigma,\varphi\ltluntil\psi} = \mprog\paren{\sigma,\psi} \vee (\mprog\paren{\sigma,\varphi} \wedge \varphi\ltluntil\psi)$
\end{itemize}
We also overload $\mprog\paren{\bsigma_{0:k},\varphi}$ to indicate multistep progression over $\bsigma_{0:k}$. To facilitate effective LTL learning on a dataset of trajectories, we restrict our consideration to LTL formulas that can be satisfied or falsified within a finite number of steps. This categories includes co-safe LTLs~\cite{6942756,lacerda2015optimal,kupferman2001model} and finite LTLs~\cite{Camacho_Baier_Muise_McIlraith_2018}. In this work, we focus on the former, which allows for evaluation over arbitrarily long sequences.

\subsection{Hierarchical RL with options}
\label{sec:pre_option}

We formulate RL within the framework of Markov decision processes (MDPs)~\cite{1994MDP}. An
MDP $M = \paren{\mathcal{S},\mathcal{A},p,r,\gamma}$
consists of
state space $\mathcal{S}$,
action space $\mathcal{A}$,
transition probability function $p:\mathcal{S}\times\mathcal{A}\rightarrow\mathscr{P}\paren{\mathcal{S}}$\footnote{$\mathscr{P}\paren{\mathcal{S}}$ defines the space of probability distributions over a set $\mathcal{S}$.},
reward function $r:\mathcal{S}\times\mathcal{A}\rightarrow\mathbb{R}$
and
discount factor $\gamma\in[0, 1)$.
Given a
policy $\pi:\mathcal{S}\rightarrow\mathscr{P}\paren{\mathcal{A}}$
and an
initial state $\mathbf{s}_0\sim\mu_0\in\mathscr{P}\paren{\mathcal{S}}$,
an agent generates a
trajectory $\btau{} = \paren{\mathbf{s}_t, \mathbf{a}_t}_{t=0}^{\infty}$
based on the distributions specified by $\pi$ and $p$. The agent receives a
discounted return $R\paren{\btau{}} = \sum_{t=0}^{\infty}\gamma^t r\paren{\mathbf{s}_t,\mathbf{a}_t}$
along its trajectory and aims to maximize the
expected return $\expect{\mu_0,\pi,p}{R\paren{\btau{}}}$.
The action-value function $Q^{\pi}\paren{\mathbf{s},\mathbf{a}}=\expect{\pi,p}{R\paren{\btau{}}\vert \mathbf{s}_0=\mathbf{s}, \mathbf{a}_0=\mathbf{a}}$
represents this expected return conditioned on a specific initial $\mathbf{s}$ and $\mathbf{a}$.
In RL, $p$ and $r$ are unknown~\cite{sutton2018reinforcement} to the learner.

The option-based framework incorporates temporally extended behaviors termed \emph{options} into MDPs~\cite{SUTTON1999181}. An
option $o=\langle \mathcal{I},\pi_o,\mathcal{B}\rangle$
is available to the agent in $\mathbf{s}$ iff $\mathbf{s}$ belongs to the
initialization set $\mathcal{I}\subseteq\mathcal{S}$.
Once the agent takes $o$, it follows $\pi_o$ until $o$ terminates, which occurs after $t$ steps according to the
probability $\mathcal{B}:\mathcal{S}^t\rightarrow[0,1]$.
Policies are defined over both primitive actions $\mathcal{A}$ and options $\mathcal{O}$, \ie,
$\pi:\mathcal{S}\rightarrow\mathscr{P}\paren{\mathcal{A}\cup\mathcal{O}}$.
The corresponding
option-value function is $Q^{\pi}\paren{\mathbf{s},o}=\expect{o,\pi,p}{R\paren{\btau{}} \vert \mathcal{E}\paren{o\pi,\mathbf{s}}}$,
where $\mathcal{E}\paren{o\pi,\mathbf{s}}$ denotes the event that $o$ is initiated in $\mathbf{s}$ at $t=0$, terminates at $t$ with probability $\mathcal{B}\paren{\mathbf{s}_1,\dots,\mathbf{s}_t}$, and  $\pi$ is followed after $o$ terminates.

\subsection{Diffusion-based Policies}

Existing diffusion-based policies directly generate a
finite trajectory $\btau{}_{0:k} = \paren{\mathbf{s}_t, \mathbf{a}_t}_{t\in\bracket{k+1}}$\footnote{The subscript $0:k$ indicates that a vector has $k+1$ elements, indexed by integers $0,1,\dots,k$. $\bracket{N}$ denotes the set of integers $\bracelr{0,1,\dots,N-1}$.}
by training on a pre-collected dataset of trajectories. These policies can be trained using expert demonstrations for imitation learning~\cite{chi2023diffusionpolicy}, or with offline datasets guided by a $Q$ function for policy improvement~\cite{janner2022diffuser}. Formally, a step-dependent neural network $s_\theta$ is trained to approximate the
score function $\nabla_{\btau{i}_{0:k}} \log \prob{i}{\btau{i}_{0:k}}$
with denoising score matching objective~\cite{vincent2011connection}:
\begin{equation}
\label{eq:dsm}
    \min_\theta \expect{i,\btau{i}_{0:k},\btau{0}_{0:k}}{\left\lVert s_\theta\left(\btau{i}_{0:k},i\right) - \nabla_{\btau{i}_{0:k}}\log\cprob{}{\btau{i}_{0:k}}{\btau{0}_{0:k}}\right\rVert^2},
\end{equation}
in which $\btau{i}_{0:k}\sim \cprob{}{\btau{i}_{0:k}}{\btau{0}_{0:k}}$ is the
data trajectory $\btau{0}_{0:k} \sim \prob{0}{\btau{0}_{0:k}}$
corrupted with noise by an $N$-step discrete approximation of
forward diffusion process $\cprob{}{\btau{i}_{0:k}}{\btau{i-1}_{0:k}}$
and $i\sim\mathcal{U}\{1,2,\ldots,N\}$.
For example, in Denoising Diffusion Probabilistic Models (DDPM)~\cite{ho2020denoising},
$\cprob{}{\btau{i}_{0:k}}{\btau{0}_{0:k}} = \mathcal{N}\left(\sqrt{\bar\alpha_i}\btau{0}_{0:k},\left(1-\bar\alpha_i\right)\bm{I}\right)$,
$\bar\alpha_i \defeq \prod_{j=1}^i \alpha_i$,
$\alpha_i \defeq 1-\beta_i$ and
$\{\beta_i\}$ is a sequence of positive noise scales $0<\beta_1,\beta_2,\dots,\beta_N<1$.
Diffusion models can generate plans conditioned on a start state by inpainting~\cite{janner2022diffuser} or classifier-free guidance~\cite{ho2021classifierfree}.

%% file: sections/method.tex
\section{Method: \method}

In this section, we introduce \method, our offline hierarchical reinforcement learning framework for Linear Temporal Logic (LTL) instructions. At a high level, \method\ integrates hierarchical RL with diffusion models to effectively address the challenges of planning under LTL constraints in an offline setting (see Figure~\ref{fig:illustration_framework}).

We begin by presenting our offline hierarchical RL approach for LTL instructions. This includes the construction of a product MDP to handle the non-Markovian nature of LTL rewards, the definition of options and rewards, and the learning of the option-value function from an offline dataset. Next, we describe how we represent options using diffusion models and introduce a novel diverse sampling technique. This approach allows us to obtain a rich set of options that are both expressive and within the support of the offline dataset, ensuring effective policy regularization.

\subsection{Offline Hierarchical RL for LTL Instructions}

Recall that we adopt an options framework for hierarchical reinforcement learning and our agent is tasked with generating behavior that satisfies a given Linear Temporal Logic (LTL) formula $\varphi \in \Psi$. The first challenge is that LTL satisfaction is evaluated over an entire trajectory. As such, defining a standard Markov Decision Process (MDP) $M$ over the native states $\mathcal{S}$ is problematic since a non-Markovian reward function cannot be defined over single states.

To address this issue, we construct a product MDP $M_{\Psi}$ such that the optimal Markov policies in $M_{\Psi}$ recover the optimal policies for the non-Markovian reward function in $M$~\cite{10.5555/3237383.3237452,pmlr-v139-vaezipoor21a}. More formally, we define $M_{\Psi} = (\mathcal{S}_{\Psi}, \mathcal{A}, p_{\Psi}, r_{\Psi}, \gamma)$, where $\mathcal{S}_{\Psi} = \mathcal{S} \times \Psi$. We assume that the properties or events related to a temporal task can be detected from the environmental state/action information and define a labeling function $L: \mathcal{S} \times \mathcal{A} \rightarrow \{0,1\}^{|\mathcal{P}|}$ that maps each state-action pair $(\mathbf{s}_t, \mathbf{a}_t)$ to a truth assignment $\sigma_t = L(\mathbf{s}_t, \mathbf{a}_t)$. We consider LTL progression~\cite{BACCHUS2000123,10.5555/3237383.3237452} as part of the transitions, and thus, the transition probability in $M_{\Psi}$ is defined as $p_{\Psi}\left(\langle \mathbf{s}', \varphi' \rangle \vert \langle \mathbf{s}, \varphi \rangle, a\right) = p(\mathbf{s}' \vert \mathbf{s}, a)$ if $\varphi' = \mprog\left(L(\mathbf{s}, \mathbf{a}), \varphi\right)$ (and 0 otherwise).  

\parabf{Options and Rewards.} We incorporate options into $M_{\Psi}$ as described in Section~\ref{sec:pre_option}. We  defer the detailed option definition to the next section and it suffices to assume that each option is a trajectory $o = \btau{}_{0:k} = \left( \mathbf{a}_t, \mathbf{s}_{t+1} \right)_{t=0}^{k-1}$. At state $\mathbf{s}_0$, the agent follows option $o$, which terminates after $k$ steps. We assume that the options have an initiation set $\mathcal{I}$ covering the entire state space.

Next, we develop an option reward function such that the agent can either (i) accomplish the instructed LTL formula $\varphi$ within the $k$ steps of an option or (ii)  transition to a state that enables completion of $\varphi$ in the future. In our product MDP $M_{\Psi}$, the agent receives a reward at $(\mathbf{s}_t, \varphi)$ given by:
\begin{equation}
    r_{\Psi}\paren{\mathbf{s}_t,\varphi,\mathbf{a}_t} = 
    \begin{cases} 
      1 & \text{if } \mprog\paren{\sigma_t,\varphi}=\true \text{ and } \varphi\neq\true\\
      -1 & \text{if } \mprog\paren{\sigma_t,\varphi}=\false \text{ and } \varphi\neq\false\\
      0 & \text{otherwise}.
    \end{cases}
\end{equation}
where $\sigma_t = L(\mathbf{s}_t, \mathbf{a}_t)$. The option reward is then defined as
$r_{\Psi}\paren{\mathbf{s}_0,\varphi,o} = \sum_{t=0}^{k-1}\gamma^t r_t$,
where $r_t = r_{\Psi}\paren{\mathbf{s}_t,\varphi_t,\mathbf{a}_t}$ and
$\varphi_t = \mprog\paren{\bsigma_{0:t},\varphi}$.

\begin{algorithm}[t]
    \small
    \caption{Offline Hierarchical RL for LTL objectives}
    \label{alg:option_rl_ltl}
    \begin{algorithmic}[1]
        \Require Datasets $\Dtraj = \big\{\btau{(d)}\big\}$ and $\Dphi$, Diffusion Model $\Diffusion{o}$. 
        \State Initialize critics $Q_{\phi_1}$, $Q_{\phi_2}$ and targets $Q_{\phi_1^{\prime}}$, $Q_{\phi_2^{\prime}}$
        \For{$e = 0$ to $E$}
            \State Sample $\big\{\btau{\paren{b}}\big\}_{b\in\bracket{B}}$ uniformly from $\Dtraj$
            \State Sample $\big\{\varphi^{\paren{b}}\big\}_{b\in\bracket{B}}$ uniformly from $\text{cl}\paren{\Dphi}$
            \LineComment{\textit{Execute following in parallel for all $b\in\bracket{B}$.}}
            \State Construct transition $\paren{\mathbf{s}_0^{\paren{b}},\varphi^{\paren{b}},o^{\paren{b}},r_{\Psi}^{\paren{b}},\mathbf{s}_k^{\paren{b}},\varphi_k^{\paren{b}}}$
            \State Propose $M$ target options ${o^{\prime}}^{(b)}_{(m)} \sim \widehat{p}_{\theta}\paren{\cdot\vert\mathbf{s}_k^{(b)}}$
            \State Select best ${o^{\prime}}^{(b)}$ by $Q_{\phi_1^{\prime}}$
            \State Get clipped Gaussian noised version $\tilde{o}^{\prime(b)}$
            \State Get loss $\ell_j^{(b)}$ by Eq.~\eqref{eq:bellman_error} with double $Q$-learning
            \State Update $\phi_j \leftarrow \phi_j - \eta\nabla_{\phi_j} \frac{1}{B}\sum_b \ell_j^{(b)}$, $j\in\bracelr{1,2}$
            \If{$e$ mod $e_0$}
                \State $\phi_j^{\prime} = \lambda\phi_j + \paren{1-\lambda}\phi_j^{\prime}$, $j\in\bracelr{1,2}$
            \EndIf
        \EndFor
        \State {\bfseries return} $Q_{\phi_1}$
    \end{algorithmic}
\end{algorithm}
    
\parabf{Learning an Option Critic.} With the above definitions, we can now proceed to learn an option-value function using an offline dataset. A high-level summary of our learning algorithm is shown Algorithm~\ref{alg:option_rl_ltl}.
According to the Bellman equation, the optimal option-value function has the form:
\begin{equation}
\label{eq:bellman_q_o}
    \begin{split}
        Q^{\ast}\paren{\langle\mathbf{s},\varphi\rangle,o}& 
        =  r_{\Psi}\paren{\mathbf{s},\varphi,o}  + \\
        & \gamma^k\sum_{\mathbf{s}^{\prime},\varphi^{\prime}} p\paren{\mathbf{s}^{\prime},\varphi^{\prime} \vert \mathbf{s},o} \max_{o^{\prime}} Q^{\ast}\paren{\langle\mathbf{s}^{\prime},\varphi^{\prime}\rangle,o^{\prime}}
    \end{split}
\end{equation}
where
$r_{\Psi}\paren{\mathbf{s},\varphi,o} = \expect{p_{\Psi}}{r_0+\cdots+\gamma^{k-1}r_{k-1}}$
and
$p\paren{\mathbf{s}^{\prime},\varphi^{\prime} \vert \mathbf{s},o}$ is the probability of transitioning to $\mathbf{s}^{\prime}$ and $\varphi^{\prime}$ when executing  option $o$ under $p_{\Psi}$. 
We aim to obtain an option critic $Q\paren{\langle\mathbf{s},\varphi\rangle,o}$ that approximates $Q^\ast$. In our offline setting, we are provided with a dataset of \emph{non-expert} trajectories $\Dtraj$ and a non-exhaustive dataset of LTL specifications $\Dphi$; in our experiments, we use tasks from prior work~\cite{pmlr-v139-vaezipoor21a,10637680}. 

We represent $Q$ using a neural network that takes in  state-option pairs and LTL formula embeddings. LTL embeddings can be obtained by using Graph Neural Networks (GNNs) that process graph representations of LTLs~\cite{1555942,4700287,Xie2021}. In this work, the LTL formula embedding is computed using a Relational Graph Convolutional Network (R-GCN)~\cite{10.1007/978-3-319-93417-4_38}, which we found performs well on new LTL formulae with the same template structure seen during training. 

We train $Q$ by minimizing a temporal difference loss, 
\begin{equation}
\label{eq:bellman_error}
    \big(r_{\Psi}\paren{\mathbf{s}_0,\varphi,o} + \gamma^k \max_{o_k\in\mathcal{O}_{\mathbf{s}_k}} Q\paren{\langle\mathbf{s}_k,\varphi_k\rangle,o_k} - Q\paren{\langle\mathbf{s}_0,\varphi\rangle,o}\big)^2
\end{equation}
using transition tuples $\paren{\mathbf{s}_0,\varphi,o,r_{\Psi},\mathbf{s}_k,\varphi_k}$ constructed using the dataset, the option set $\mathcal{O}_{\mathbf{s}_k}$, and the training LTL space. Specifically, we sample a \emph{short-horizon}
trajectory $\btau{}_{0:k} \sim \prob{0}{\btau{}_{0:k}}$
 from the dataset. Each $\btau{}_{0:k} = \paren{\mathbf{s}_t, \mathbf{a}_t}_{t=0}^k$ includes
current state $\mathbf{s}_0$, option $o=\paren{\mathbf{a}_t,\mathbf{s}_{t+1}}_{t=0}^{k-1}$  and next state $\mathbf{s}_k$.
We use $L$ to obtain a truth assignment sequence $\bsigma_{0:k}$ and compute the reward $r_{\Psi}\paren{\mathbf{s}_0,\varphi,o} = \sum_{t=0}^{k-1}\gamma^t r_t$. The next LTL $\varphi_k$ is obtained via LTL progression. We construct the training LTL space from the progression closure of formulas $\varphi \in \Dphi$, denoted $\text{cl}\paren{\Dphi}$. The loss function Eq.~(\ref{eq:bellman_error}) is trained by uniformly sampling trajectories from dataset and LTL instructions from $\text{cl}\paren{\Dphi}$.

To stabilize the training, we leverage several popular techniques for $Q$ networks. This includes using time delayed version of $Q$~\cite{pmlr-v48-mniha16}, clipped double $Q$-Learning~\cite{NIPS2010_091d584f,pmlr-v80-fujimoto18a}, and smoothing the target policy by injecting noise into the target actions~\cite{simmons-edler2019qlearning,pmlr-v80-fujimoto18a}. 

\subsection{Options with Diffusion}
Thus far, we have not yet fully defined our option set. Ideally, we would like a rich option set that is expressive enough to cover the diverse requirements of LTL instructions. However, one crucial issue is that in the offline setting, the dataset available to the learner is fixed and exploration is not permitted. As such, the options should generate trajectories within the support of the offline dataset to ensure reliable Q-value estimation.

To address this problem, we propose to regularize the policy space using diffusion options, i.e., we represent our option set $\mathcal{O}$ using a diffusion model and limit the policy's co-domain to be $\mathcal{O}$ (instead of $\mathcal{A}\cup\mathcal{O}$). Diffusion models form a rich policy class that can capture multimodal distributions, while generating samples within the training dataset. 

Once a diffusion model $\Diffusion{\btau{}_{0:k}}$ is trained on the dataset, it can be used to sample options $o = \btau{}_{0:k} \sim \Diffusion{\btau{}_{0:k}}$.
However, during both training (Eq.~\ref{eq:bellman_error}) and deployment, we need to search over options to maximize the $Q$ function. With diffusion models, finding the optimal option by sampling may require a large sample size to reduce approximation error. In addition, the samples may lack diversity, with many similar trajectories from a few modes of the distribution.

\parabf{Q-Guidance.} One potential way forward is to leverage the trained $Q$-network to perform conditional sampling. This form of classifier guidance can be achieved by methods such as posterior sampling. However, we find that in practice, $Q$-guided generation tends to produce low-quality trajectories (see Section~\ref{sec:ablation}).

\parabf{Diversity Sampling.} We propose an alternative sampling approach that covers different modes of data distribution, while keeping the sample batch size $M$ as small as possible. 

The key idea is that each generated trajectory should be distinct from the others in the batch. A standard diffusion model generates options using an approximate conditional score $s_{\theta}\paren{\cdot\vert\mathbf{s}_0} \approx \nabla_{\btau{i}_{0:k}}\log\cprob{i}{\btau{i}_{0:k}}{\mathbf{s}_0}$. We instead propose to sample from a posterior that conditions on the $M-1$ other samples with the
conditional score function $\nabla_{\btau{i}_{\paren{1}}}\log \cprob{i}{\btau{i}_{\paren{1}}}{\bracelr{\hat{\btau{}}_{\paren{m}}}_{m=2,\dots,M}}$.
In the following text we omit the subscripts $0:k$ and conditioned state $\mathbf{s}_0$ for clarity.

With this change, the $M$ samples are no longer independent. We leverage Bayes' rule and sample estimation~\cite{chung2023diffusion} to express the conditional score as the sum of two terms:
$\nabla_{\btau{i}_{\paren{1}}}\log p_{i}\big(\btau{i}_{\paren{1}}\big)$
and
$\nabla_{\btau{i}_{\paren{1}}}\log\ p\big(\bracelr{\hat{\btau{}}_{\paren{m}}}_{m=2,\dots,M} \vert \hat{\btau{}}_{\paren{1}}\big)$,
where the noiseless trajectory $\hat{\btau{}}_{\paren{1}}$ is estimated via
Tweedie’s formula~\cite{efron2011tweedie}
$\hat{\btau{}}_{\paren{1}} = \frac{1}{\sqrt{\bar\alpha_i}} \big(\btau{i}_{\paren{1}} + \paren{1-\bar\alpha_i} \nabla_{\btau{i}_{\paren{1}}} \log p_{i}\big(\btau{i}_{\paren{1}}\big)\big)$
and
$p_{i}\big(\{\hat{\btau{}}_{\paren{m}}\}_{m=2,\dots,M} \mid \btau{i}_{\paren{1}}\big)$
is approximated by point estimation
$p\big(\{\hat{\btau{}}_{\paren{m}}\}_{m=2,\dots,M} \mid \hat{\btau{}}_{\paren{1}}\big)$.

We model this conditional probability using a differentiable probabilistic model. To encourage sample diversity, the conditional probability should be high if $\hat{\btau{}}_{\paren{1}},\dots,\hat{\btau{}}_{\paren{M}}$ are pairwise dissimilar, and low if any of them are similar. 
Thus, the unnormalized probability of the generated set can be modeled by a similarity matrix analogous to determinantal point processes~\cite{kulesza2012determinantal}, 
\begin{equation}
\cprob{}{\hat{\btau{}}_{\paren{1}}}{\bracelr{\hat{\btau{}}_{\paren{m}}}_{m=2}^{M}} = \prob{}{\hat{\btau{}}_{\paren{1}}} \det\paren{L_M}/Z,
\end{equation}
where $L_M$ is the similarity matrix with elements indexed by integers $(u,v)$, $1\leq u,v\leq M$ measuring the similarity (\eg, cosine similarity)
between $\hat{\btau{}}_{\paren{u}}$ and $\hat{\btau{}}_{\paren{v}}$, and
$Z = \int_{\hat{\btau{}}_{\paren{1}}} \prob{}{\hat{\btau{}}_{\paren{1}}} \det\paren{L_M} \diff\hat{\btau{}}_{\paren{1}}$ is a normalizing constant~\cite{pmlr-v202-song23k}.
Since the normalizing constant $Z$ does not depend on $\hat{\btau{}}_{\paren{1}}$, the gradient of the point estimate becomes
\begin{equation}
    \nabla_{\btau{i}_{\paren{1}}} \log\cprob{}{\bracelr{\hat{\btau{}}_{\paren{m}}}_{m=2}^{M}}{\hat{\btau{}}_{\paren{1}}} = \nabla_{\btau{i}_{\paren{1}}} \log\det\paren{L_M}.
\end{equation}
This gradient can be plugged into the reverse process for posterior sampling as summarized in Algorithm~\ref{alg:diversity_ps}. Our approach maximizes the determinant of the similarity matrix, biasing the generated trajectories to be distinct from one another, thus promoting diversity.

\begin{algorithm}
    \small
    \caption{Diversity Guided Batch Sampling}
    \label{alg:diversity_ps}
    \begin{algorithmic}[1]
        \Require $N$, $s_{\theta}$, $M$, $\{\zeta_i\}_{i=1}^N$
        \LineComment{\textit{Execute following in parallel for all $m\in\bracket{M}$.}}
        \State $\btau{N}_{(m)} \sim \mathcal{N}\left(\bm{0}, \bm{I}\right)$, $m\in\bracket{M}$
        \For{$i = N-1$ {\bfseries to} $0$}
            \State{{$\hat{s}^{(m)} \gets s_{\theta}\left(\btau{i}_{(m)},i\right)$}}
            \State Predict $\hat{\bm{\tau}}^{0}_{(m)}$ via Tweedie’s formula
            \State Reverse one step to get $\btau{i-1}_{(m)}$
            \State $L_{u,v} = \mathrm{cos}\paren{\hat{\btau{}}^{0}_{(u)},\hat{\btau{}}^{0}_{(v)}}$, $u,v\in\bracket{M}$
            \State{$\btau{i-1}_{(m)} \gets \btau{i-1}_{(m)} + {\zeta_i} \nabla_{\btau{i}_{(m)}} \log\det\paren{L}$}
        \EndFor
        \State {\bfseries return} $\bracelr{\hat{\bm{\tau}}^{0}_{(m)}}_{m\in\bracket{M}}$
    \end{algorithmic}
\end{algorithm}
\vspace{-10pt}

%% file: sections/relatedwork.tex
\section{Related Work}

LTL is widely-used to specify high-level, temporally extended requirements in robot tasks~\cite{BACCHUS2000123,10.5555/3037104.3037150,1570410}. 
Many existing data-driven methods for LTL planning learn in an \emph{online} fashion~\cite{ijcai2022p507,10.5555/3237383.3237452,voloshin2022policy,pmlr-v139-vaezipoor21a,TIAN2023104351}. These methods can be effective but may also lead to unsafe interactions during the trial-and-error learning process. 

Here, we focus on offline learning that do not require interactions with the environment. Existing offline methods typically rely on knowledge of the environmental dynamics, are limited to certain types of LTL specifications~\cite{5399536,10611432}, or require events to be explicitly labeled in the dataset~\cite{wang2022temporal}. The closest related work is our previous model \ltldog{}~\cite{10637680}, which uses posterior sampling with diffusion models and LTL model checkers to generate plans under LTL instructions. However, \ltldog{} is restricted to non-hierarchical open-loop planning. Subgoal generation via hierarchical diffusion frameworks has been explored using methods such as graph search~\cite{pmlr-v202-li23ad} and keyframe discovery~\cite{ma2024hierarchical}. However, these methods are not designed for satisfying LTL specifications. To our knowledge, we are the first to successfully combine offline hierarchical RL with diffusion-based options for LTL planning, enabling effective closed-loop control under complex temporal specifications. \method{} learns \emph{without} access to task-oriented expert trajectories and as experiments will show, can produce effective long-horizon trajectories to comply with novel LTL instructions.

%% file: sections/experiments.tex
\begin{figure*}
    \centering
    \subfigure[Medium]{
        \includegraphics[width=0.26\columnwidth]{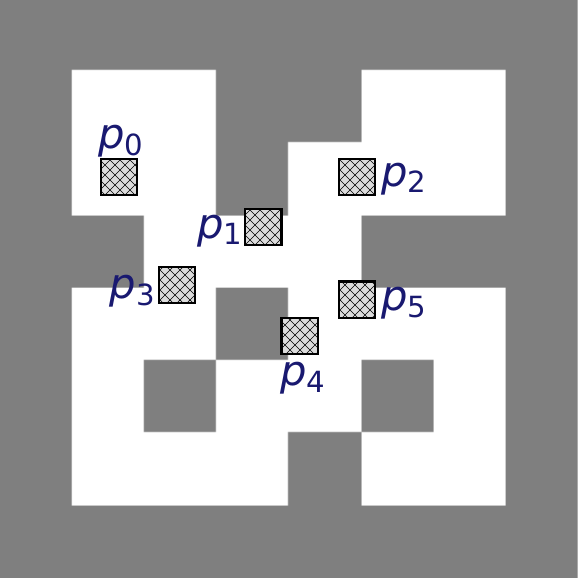}
        \label{fig:illustration_maze2d_medium}
    }
    \hfill
    \subfigure[\ltldog]{
        \includegraphics[width=0.26\columnwidth]{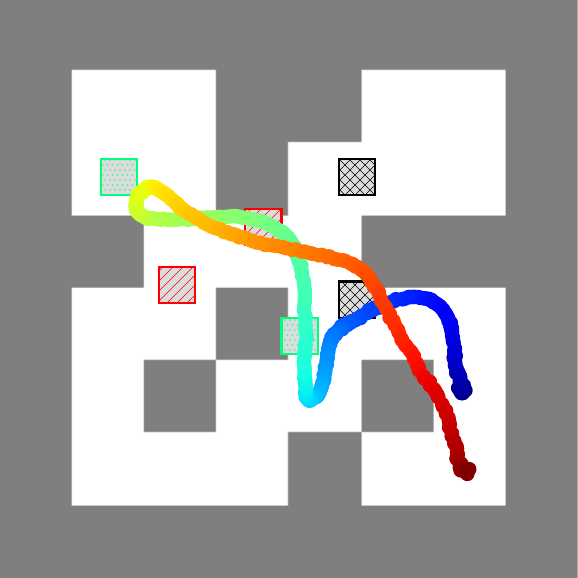}
        \label{fig:ltl_maze2d_medium_dps}
    }
    \hfill
    \subfigure[\method]{
        \includegraphics[width=0.26\columnwidth]{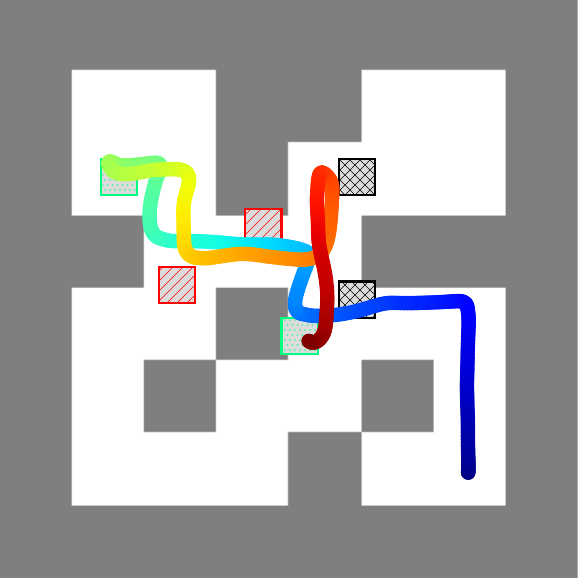}
        \label{fig:ltl_maze2d_medium_rldpp}
    }
    \hfill
    \subfigure[Large]{
        \includegraphics[width=0.34\columnwidth]{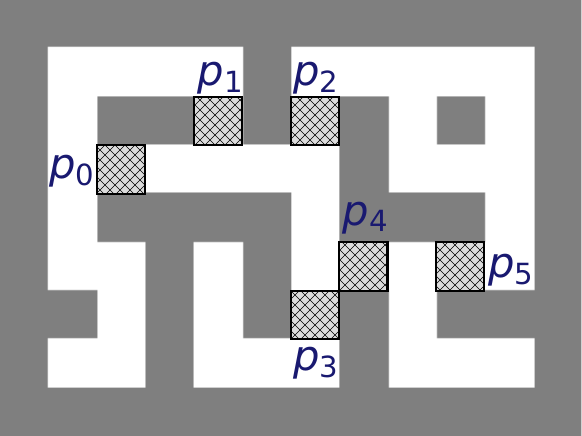}
        \label{fig:illustration_maze2d_large}
    }
    \hfill
    \subfigure[\ltldog]{
        \includegraphics[width=0.34\columnwidth]{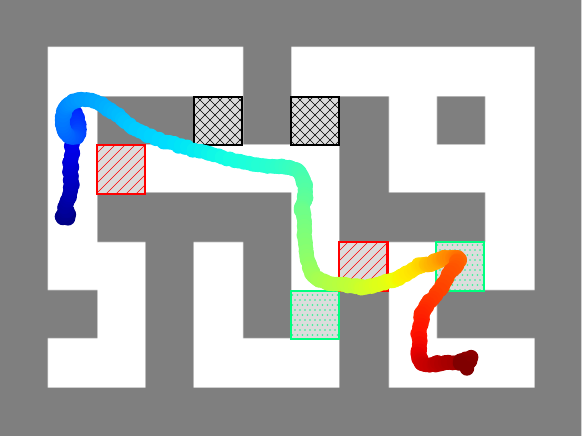}
        \label{fig:ltl_maze2d_large_dps}
    }
    \hfill
    \subfigure[\method]{
        \includegraphics[width=0.34\columnwidth]{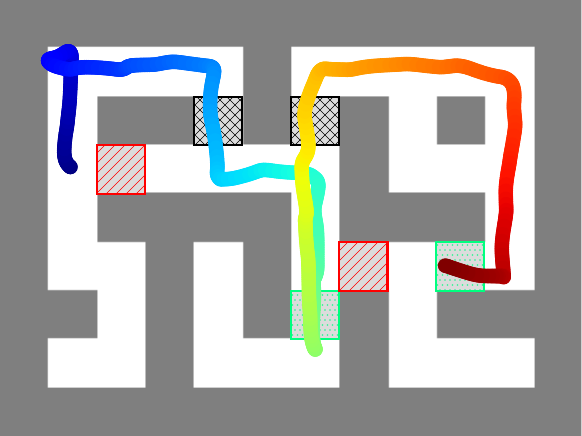}
        \label{fig:ltl_maze2d_large_rldpp}
    }
    \vspace*{-5pt}
    \caption{
    Setup and sample trajectories for Maze2D environments. (a) and (d) depict Maze2D Medium and Large, each containing six non-overlapping regions (hatched squares labeled with $p_x$) used to evaluate atomic propositions in $\mathcal{P}$. The agent is tasked with visiting specific regions in different temporally extended sequences. (b) and (e) illustrate the trajectories generated by \ltldog and our proposed method (\method) (from blue to red) under the specification $\varphi = \neg p_3 \ltluntil (p_0 \wedge (\neg p_1 \ltluntil p_4))$. (c) and (f) show trajectories generated under the specification $\varphi = \neg p_0 \ltluntil (p_3 \wedge (\neg p_4 \ltluntil p_5))$. Our method (\method) successfully satisfies these complex LTL specifications by avoiding regions with $\neg$ propositions (red zones) before reaching the designated green regions, as demonstrated in panels (e) and (f).
    }
    \label{fig:illustration_maze2d}
    \vspace{-5pt}
\end{figure*}

\begin{figure}
    \centering
    \subfigure[PushT Setup]{
        \includegraphics[width=.29\columnwidth]{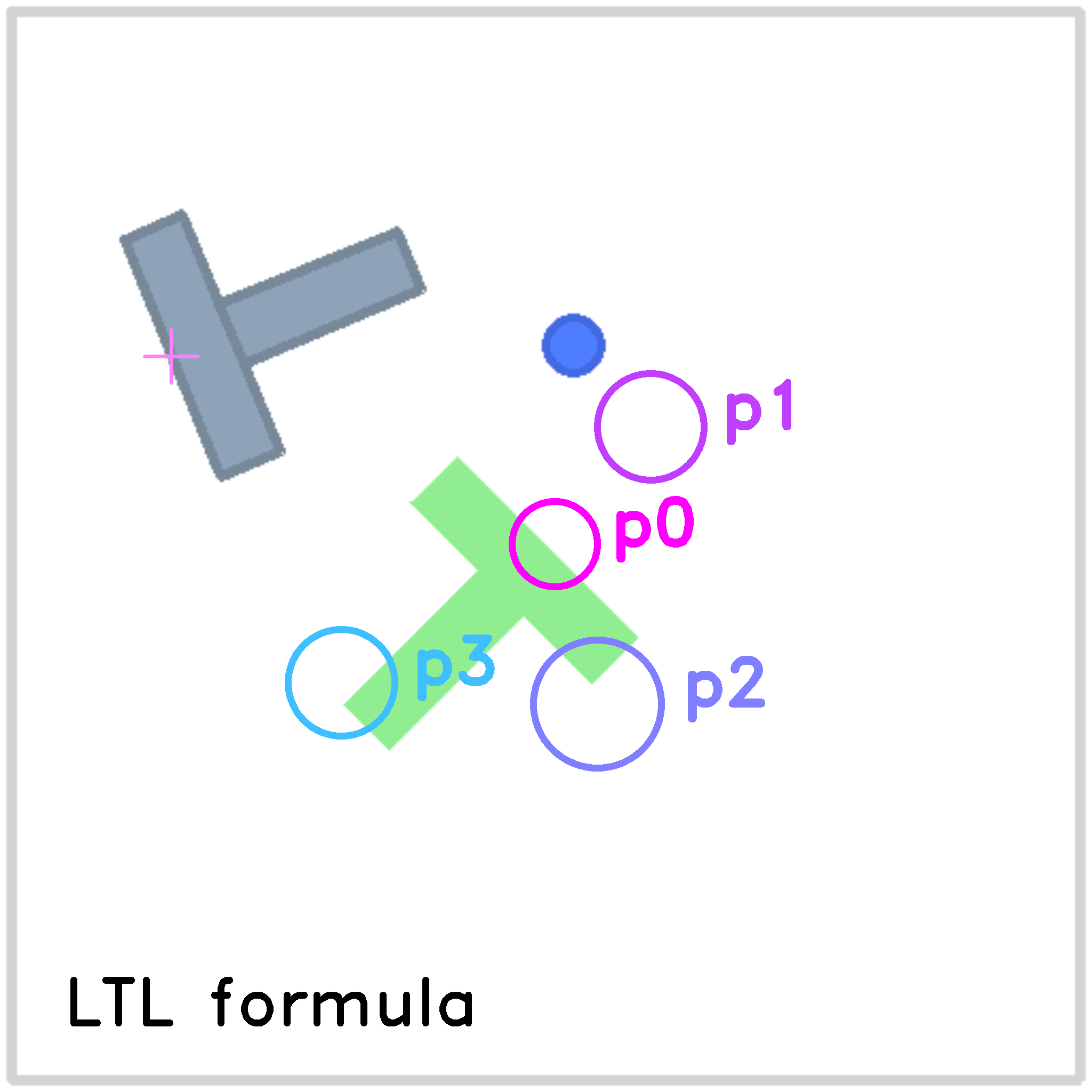}
        \label{fig:illustration_pusht_regions}
    }
    \hfill
    \subfigure[\ltldog]{
        \includegraphics[width=.29\columnwidth]{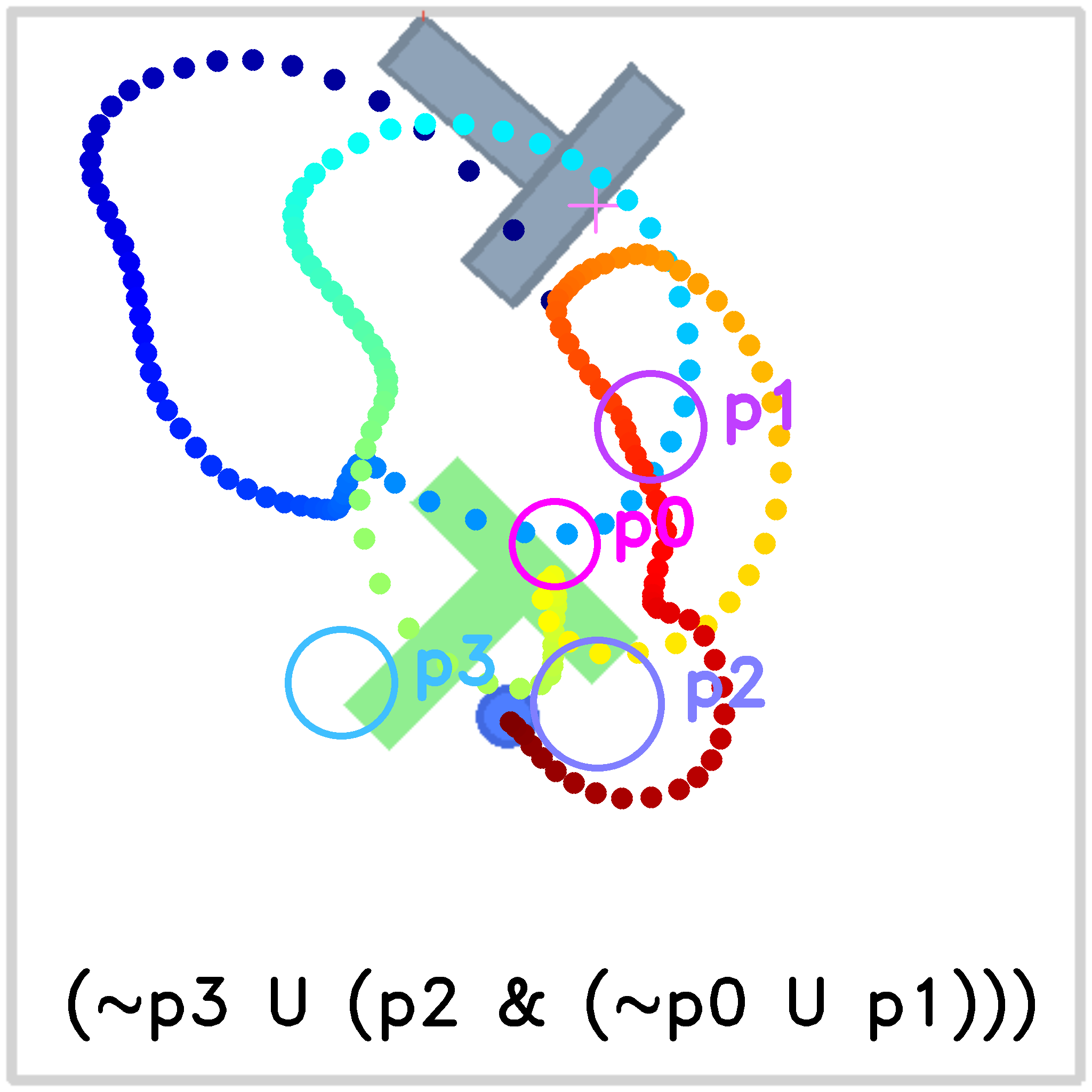}
        \label{fig:ltl_pusht_ltldog}
    }
    \hfill
    \subfigure[\method (Ours)]{
        \includegraphics[width=.29\columnwidth]{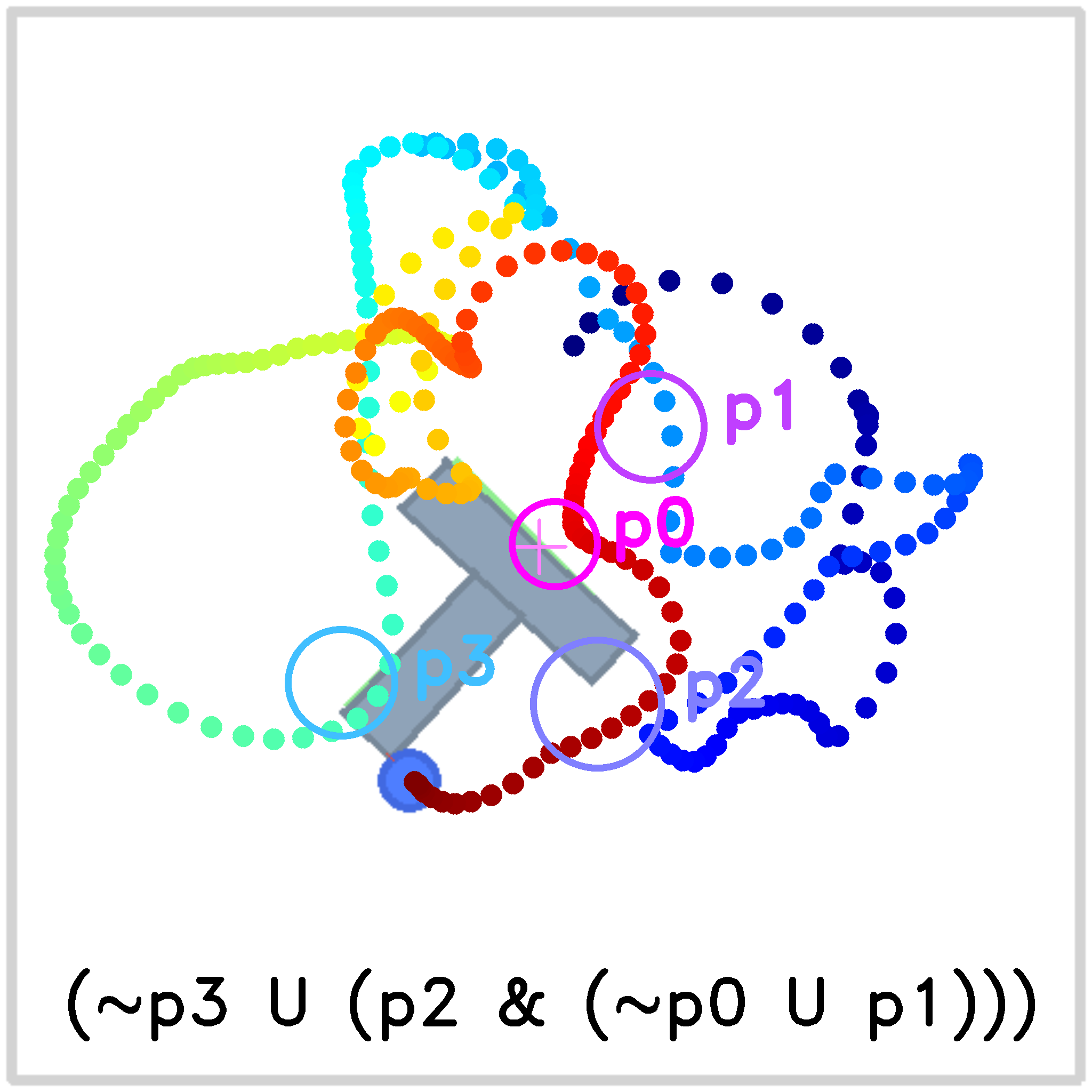}
        \label{fig:ltl_pusht_rldpp}
    }
    \vspace*{-5pt}
    \caption{
        The PushT manipulation environment. 
        (a) A robot arm's end effector (circles filled in blue) should manipulate the \texttt{T} block (gray) to a goal pose (green) \emph{and} visit some regions (hollow circles marked with $p_x$) under different temporally-extended orders before completion. 
        (b) \ltldogr does not comply with the LTL nor completes the manipulation.
        (c) In contrast, \method can satisfy the LTL \emph{and} complete the manipulation task. 
    }
    \label{fig:illustration_pusht}
    \vspace{-12pt}
\end{figure}

\begin{figure}
    \centering
    \subfigure[Studio-like Lab]{
        \includegraphics[width=.29\columnwidth]{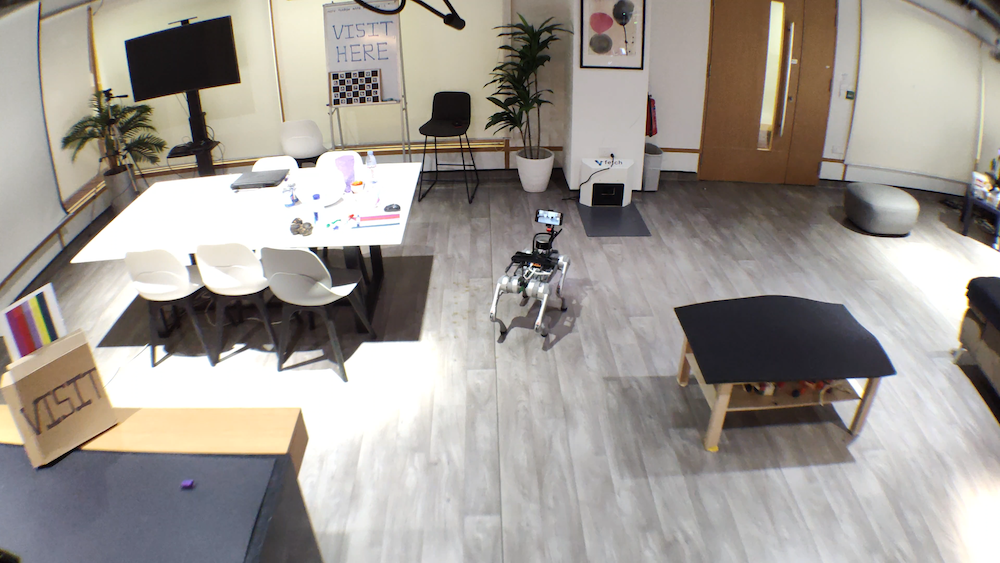}
        \label{fig:illustration_rls}
    }
    \hfill
    \subfigure[Office room]{
        \includegraphics[width=.29\columnwidth]{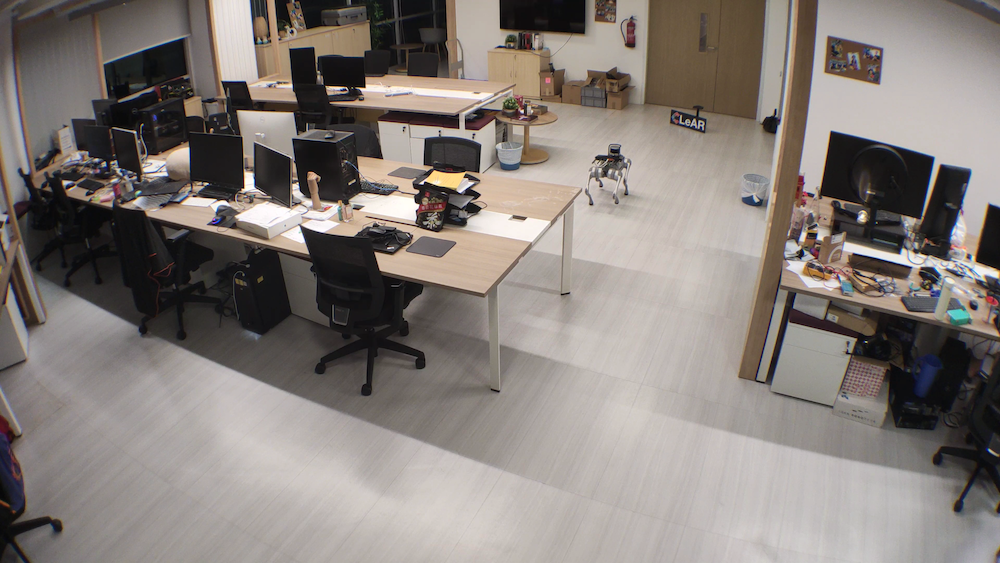}
        \label{fig:illustration_ssi}
    }
    \hfill
    \subfigure[Unitree Go2]{
        \includegraphics[width=.29\columnwidth]{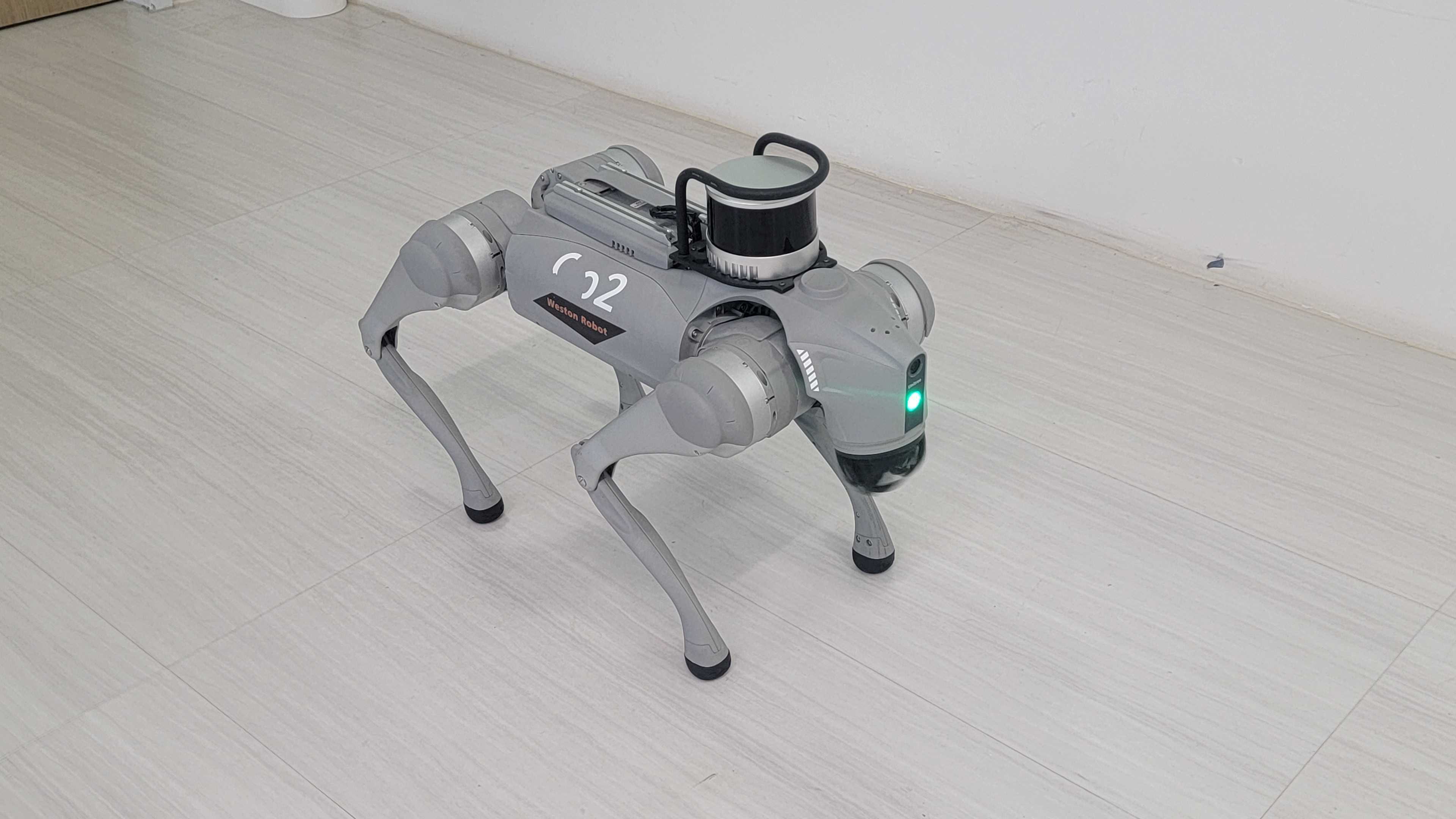}
        \label{fig:illustration_dog}
    }
    \vspace*{-5pt}
    \caption{Real world environments for quadruped robot navigation.}
    \label{fig:illustration_real}
    \vspace{-10pt}
\end{figure}

\begin{figure}
    \centering
    \subfigure[Lab]{
        \includegraphics[width=.295\columnwidth,
            trim=0 0 0 0.9cm, 
            clip
        ]{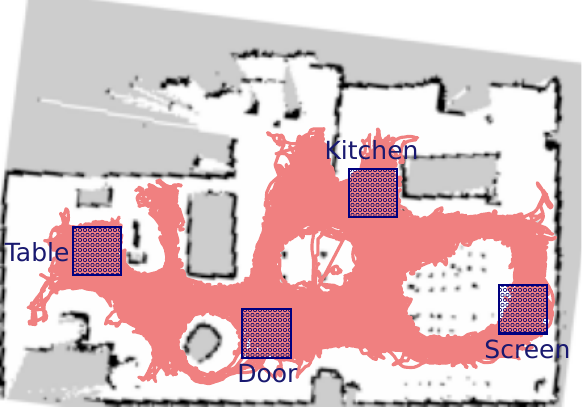}
        \label{fig:illustration_map_rls}
    }
    \hfill
    \subfigure[\ltldog]{
        \includegraphics[width=.295\columnwidth,
            trim=0 0 0 0.9cm, 
            clip
        ]{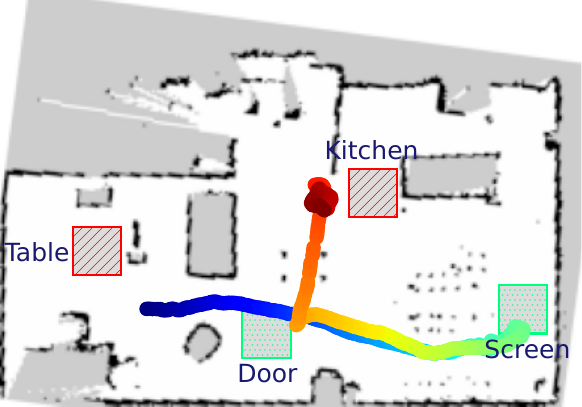}
        \label{fig:ltl_rls_dps}
    }
    \hfill
    \subfigure[Ours]{
        \includegraphics[width=.2951\columnwidth,
            trim=0 0 0 0.9cm, 
            clip
        ]{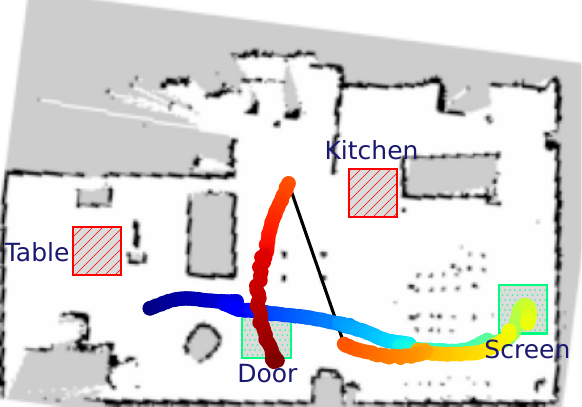}
        \label{fig:ltl_rls_rldpp}
    }
    \hfill
    \subfigure[Office]{
        \includegraphics[width=.295\columnwidth,
            trim=0.5cm 0 0.4cm 2.3cm, 
            clip
        ]{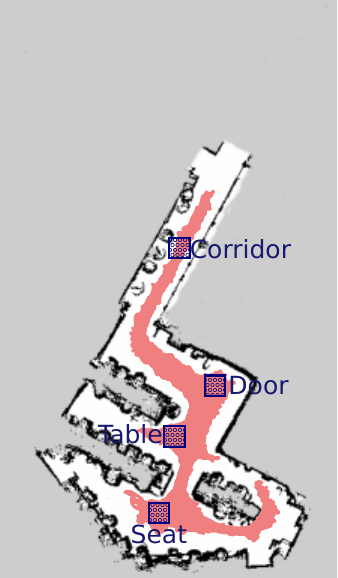}
        \label{fig:illustration_map_ssi}
    }
    \hfill
    \subfigure[\ltldog]{
        \includegraphics[width=.295\columnwidth,
            trim=0.5cm 0 0.4cm 2.3cm, 
            clip
        ]{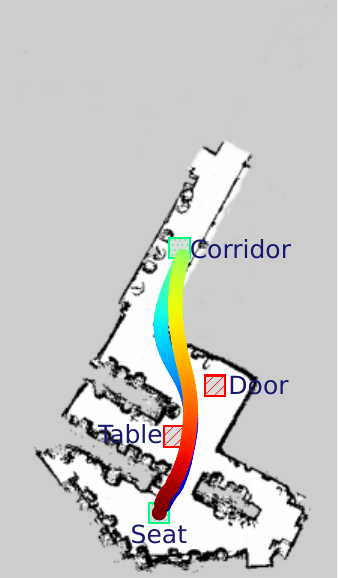}
        \label{fig:ltl_ssi_dps}
    }
    \hfill
    \subfigure[Ours]{
        \includegraphics[width=.2951\columnwidth,
            trim=0.5cm 0 0.4cm 2.3cm, 
            clip
        ]{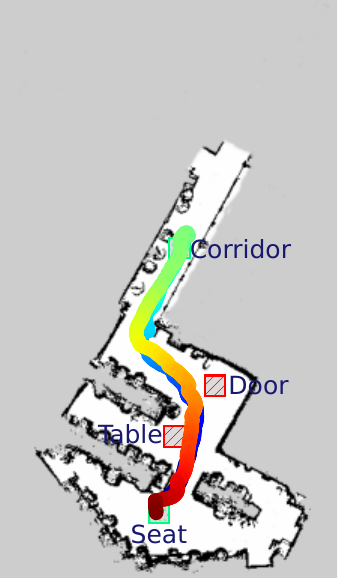}
        \label{fig:ltl_ssi_rldpp}
    }
    \vspace*{-12pt}
    \caption{Results in real-world rooms. Each room has 4 key locations ((a) and (d)).
    The instructed LTL is $\neg \text{Table} \ltluntil (\text{Screen} \wedge (\neg \text{Kitchen} \ltluntil \text{Door}))$ for lab (first row) and $\neg \text{Door} \ltluntil (\text{Corridor} \wedge (\neg \text{Table} \ltluntil \text{Seat}))$ for office (second row). \ltldog is unable to recover from perturbations (b) and cannot generate a valid plan (e), while ours ((c) and (f)) can achieve both.}
    \label{fig:result_lab_ltl}
    \vspace{-18pt}
\end{figure}

\section{Experiments}

Our experiments aim to evaluate \method's ability to learn and plan for temporal tasks that are not explicitly demonstrated in the dataset. First, we compare \method\ to state-of-the-art (SOTA) data-driven methods with temporally extended instructions in simulated benchmark environments.  Then, we demonstrate \method's robustness in dealing with control noise and external perturbations in real-world tasks through a case study on a quadruped robot (Figure~\ref{fig:illustration_real}). The section concludes with an ablation study on our diversity sampling approach.

\begin{table}
    \centering
    \begin{threeparttable}
    \caption{Satisfaction rate ($\%$) ($\uparrow$) on Different LTLs in Maze2d. }
    \label{table:maze2d_ltl}
    \renewcommand{\tabcolsep}{3pt}
    \renewcommand{\arraystretch}{0.9}
    \begin{tabular}{llccccc}
        \toprule
        \multirow{2}{*}{\textbf{Env.}} & \multirow{2}{*}{\textbf{Method}} && \multicolumn{2}{c}{Training $\ltlf$s} & \multicolumn{2}{c}{Testing $\ltlf$s} \\
        \cmidrule{4-7}
        \cmidrule{4-5} \cmidrule{6-7}
        &&& Planning & Rollout & Planning & Rollout \\
        \midrule
        \multirow{3}{*}{\textbf{Medium}}    & \diffuser && 15.0$\pm$0.7 & 13.4$\pm$0.6          & 11.6$\pm$1.4 & 10.1$\pm$1.2 \\
                                            & \ltldogs  && 77.9$\pm$5.7 & 31.8$\pm$2.6          & 68.4$\pm$6.7 & 28.7$\pm$3.5 \\
                                            & \ltldogr  && 51.8$\pm$1.8 & 39.5$\pm$1.6          & 43.3$\pm$4.4 & 30.6$\pm$1.9 \\ 
                                            & \method   && \bf{95.3$\pm$0.8} & \bf{90.2$\pm$3.3}      & \bf{95.7$\pm$0.6} & \bf{89.6$\pm$2.1} \\
                                            \midrule                                                        
        \multirow{3}{*}{\textbf{Large}}     & \diffuser && 13.5$\pm$0.4 & 12.8$\pm$0.1          & 11.6$\pm$2.3 & 11.5$\pm$1.7 \\
                                            & \ltldogs  && 73.8$\pm$2.4 & 32.6$\pm$1.4          & 66.6$\pm$2.7 & 24.9$\pm$1.7 \\
                                            & \ltldogr  && 66.9$\pm$0.6 & 47.4$\pm$0.8          & 57.5$\pm$2.3 & 39.0$\pm$2.9 \\ 
                                            & \method   && \bf{95.4$\pm$0.3} & \bf{92.0$\pm$0.5}      & \bf{94.1$\pm$1.6}	&	\bf{89.8$\pm$2.0} \\
                                            \bottomrule
    \end{tabular}
    \end{threeparttable}
    \vspace*{-5pt}
\end{table}

\begin{table}
    \centering
    \begin{threeparttable}
    \caption{Performance on Different LTLs in PushT}
    \label{table:pusht_ltl}
    \renewcommand{\tabcolsep}{3pt}
    \begin{tabular}{llcccc}
        \toprule
        \textbf{Method\textbackslash Performance} & \textbf{LTL Set} && Satisf. rate ($\%$) $\uparrow$ & Score $\uparrow$ \\
        \midrule
        \multirow{2}{*}{\diffusionpolicy}   & Training && 21.8$\pm$1.02   & 0.388$\pm$0.006 \\
                                            & Test     && 32.8$\pm$1.48   & 0.385$\pm$0.028 \\
        \multirow{2}{*}{\ltldogs}           & Training && 28.7$\pm$1.30   & 0.294$\pm$0.006 \\ 
                                            & Test     && 38.9$\pm$2.91   & 0.300$\pm$0.023 \\ 
        \multirow{2}{*}{\ltldogr}           & Training && 71.9$\pm$1.77   & 0.289$\pm$0.001 \\ 
                                            & Test     && 62.9$\pm$1.72   & 0.353$\pm$0.004 \\
        \multirow{2}{*}{\method}            & Training && \bf{87.4$\pm$0.84}  & \bf{0.386$\pm$0.022}\\ 
                                            & Test     && \bf{94.1$\pm$1.85}  & \bf{0.432$\pm$0.042} \\ 
                                            \bottomrule 
    \end{tabular}
    \end{threeparttable}
    \vspace*{-10pt}
\end{table}

\begin{table}
    \centering
    \begin{threeparttable}
    \caption{Results for Real-World Navigation Tasks.}
    \label{table:result_sr_rooms}
    \renewcommand{\tabcolsep}{5pt}
    \renewcommand{\arraystretch}{0.9}
    \begin{tabular}{llccc}
        \toprule
        \multirow{2}{*}{\textbf{Environment}} & \multirow{2}{*}{\textbf{Method\textbackslash Performance}} && \multicolumn{2}{c}{Satisfaction rate ($\%$) $\uparrow$} \\
        \cmidrule{4-5}
        &&& LTL & Perturbation \\
        \midrule
        \multirow{2}{*}{\textbf{Lab}}       & \ltldog     && 70 & 45 \\
                                            & \method     && \bf{95} & \bf{95} \\
                                            \midrule
        \multirow{2}{*}{\textbf{Office}}    & \ltldog     && 30 & 15 \\
                                            & \method     && \bf{95} & \bf{85} \\
                                            \bottomrule
    \end{tabular}
    \end{threeparttable}
    \vspace*{-15pt}
\end{table}

\begin{table*}
    \centering
    \begin{threeparttable}
    \caption{Ablation Study on Different LTLs in Maze2d-Large. }
    \label{table:ablation}
    \renewcommand{\tabcolsep}{1.7pt}
    \renewcommand{\arraystretch}{0.9}
    \begin{tabular}{lccccccccccccc}
        \toprule
        \multirow{3}{*}{\textbf{Method\textbackslash Performance}} && \multicolumn{6}{c}{Training LTLs} & \multicolumn{6}{c}{Testing LTLs} \\
        \cmidrule{3-14}
        && \multicolumn{2}{c}{Satisf. rate ($\%$) $\uparrow$} & \multicolumn{2}{c}{Failure rate ($\%$) $\downarrow$} & \multicolumn{2}{c}{Successful Steps $\downarrow$} & \multicolumn{2}{c}{Satisf. rate ($\%$) $\uparrow$} & \multicolumn{2}{c}{Failure rate ($\%$) $\downarrow$} & \multicolumn{2}{c}{Successful Steps $\downarrow$} \\
        \cmidrule{3-4} \cmidrule{5-6} \cmidrule{7-8} \cmidrule{9-10} \cmidrule{11-12} \cmidrule{13-14}
        && Planning & Rollout & Planning & Rollout & Planning & Rollout & Planning & Rollout & Planning & Rollout & Planning & Rollout\\
                                \midrule
                                
        $Q$-Guidance            && 82.4$\pm$1.0	&	80.1$\pm$0.7	&	2.6$\pm$0.5	&	\bf{2.6$\pm$0.3}	&	759.4$\pm$13.7	&	819.4$\pm$4.0	&	79.1$\pm$1.2	&	76.1$\pm$0.9	&	2.8$\pm$0.5	&	2.6$\pm$0.9	&	814.9$\pm$10.8	&    905.5$\pm$15.5 \\
        Ours (w.o. diversity) && 91.0$\pm$0.8	&	91.0$\pm$0.7	&	4.6$\pm$2.3	&	5.6$\pm$2.1	&	515.1$\pm$13.8	&	572.5$\pm$14.2	&	89.5$\pm$0.7	&	88.4$\pm$2.2	&	4.0$\pm$2.7	&	5.4$\pm$3.2	&	587.9$\pm$8.0	&	647.3$\pm$13.1 \\
        Ours (\method)                 && \bf{95.4$\pm$0.3}	&	\bf{92.0$\pm$0.5}	&	\bf{2.2$\pm$0.4}	&	2.8$\pm$0.6	&	\bf{386.9$\pm$5.2}	&	\bf{445.7$\pm$12.4}	&	\bf{94.1$\pm$1.6}	&	\bf{89.8$\pm$2.0}	&	\bf{1.8$\pm$0.7}	&	\bf{2.4$\pm$0.6}	&	\bf{429.1$\pm$10.0}	&	\bf{516.7$\pm$14.5} \\
                                \bottomrule
    \end{tabular}
    \end{threeparttable}
    \vspace*{-12pt}
\end{table*}

\subsection{Experimental Setup}

\parabf{Environments.} We conduct experiments in benchmark simulation environments ---Maze2d~\cite{janner2022diffuser,fu2020d4rl} and PushT~\cite{chi2023diffusionpolicy} --- and on a real quadruped robot. In Maze2d (Fig.~\ref{fig:illustration_maze2d}), tasks specified by LTL formulae require complex, long-horizon navigation beyond simple goal-reaching. LTL properties and events are determined by the agent's presence in key regions of the maze. We adopt the LTL setting in prior work~\cite{10637680}, where the regions of states and actions for each atomic proposition in $\mathcal{P}$ are disjoint. In the PushT task (Fig.~\ref{fig:illustration_pusht}), the agent manipulates a \texttt{T} block via a controllable mover to accomplish temporal instructions. In real-world experiments, we test LTL tasks on a quadruped robot in indoor environments (Fig.~\ref{fig:illustration_real}). To increase problem difficulty, we perturbed the robot by moving it manually to a different position during the rollout (reminiscent of the kidnapped robot problem). This was to test if the planner could recover from large deviations from the intended plan.

\parabf{Compared Methods.} We compare \method\ against other data-driven planning methods: \diffuser\ and \diffusionpolicy, which are diffusion-based methods for sampling plans but lack the ability to plan over temporal constraints. The most relevant work is \ltldog~\cite{10637680}, which generates full trajectories under LTL instructions. We evaluate performance using the average LTL satisfaction rate during planning and rollout, averaged over 10 random start locations for each LTL. In the real-robot experiment, we executed 80 trajectories in total for each method. 

\subsection{Results}

\parabf{Comparative Analysis of Methods.} Tables~\ref{table:maze2d_ltl} and \ref{table:pusht_ltl} present the performance under randomly-generated LTL specifications using the \texttt{Until} sampler, as in~\cite{10637680,pmlr-v139-vaezipoor21a} ($200$ for Maze2d and $36$ for PushT), which contain different sequences of regions to visit and avoid. We observe that \method{} achieves substantially higher success rates than all baselines. These results demonstrate that \method{} effectively discovers compositions of skills and benefits from closed-loop planning.

Illustrated examples in Figure~\ref{fig:illustration_maze2d} show that \method's closed-loop nature successfully handles long trajectories that exceed the fixed horizon on which \ltldog{} was trained. Note that some tasks may be impossible to accomplish due to physical limitations such as the agent's starting location, obstacles, or the arrangement of propositional regions in the maze, which limits performance below $100\%$. 

\parabf{Real-World Study on Quadruped Robot.} As shown in Table~\ref{table:result_sr_rooms} and Figure~\ref{fig:result_lab_ltl}, \method's outperforms \ltldog, achieving significantly higher satisfaction rates. \ltldog\ lacks replanning capabilities and was prone to task violations when perturbed to be close to avoidance regions. In contrast, \method\ exhibits much more robust behavior, with only a few failure cases; these failures typically occurred in out-of-distribution areas, where the diffusion model struggled to generate valid options.

\label{sec:ablation}
\parabf{Ablation Study on the Diversity Sampling.} Our proposed diversity sampling method significantly outperforms standard sampling and Q-guidance (see \tabref{table:ablation}). The gap between Q-guidance and the other methods is relatively large, which we posit is due to mismatched $Q$-guidance gradients with the denoising updates in diffusion.

%% file: sections/conclusion.tex
\section{Conclusion, Limitations and Future Work} 

In this paper, we introduced \method{}, an offline hierarchical reinforcement learning framework that leverages diffusion-based options to satisfy complex LTL instructions. Our approach effectively integrates diffusion models into a hierarchical RL setting, enabling the agent to generate diverse and expressive behaviors while adhering to the constraints of the offline dataset. Experiments in both simulation and real-world environments demonstrated that \method{} significantly outperforms state-of-the-art methods in satisfying temporal specifications and exhibits robustness to control noise and external perturbations.

A limitation of our method is its reliance on the availability of relevant behavioral fragments within the offline dataset. When the required behaviors cannot be constructed from fragments of trajectories in the dataset, \method{} may not to find a suitable option. Future work includes exploring techniques to overcome this limitation. 